\documentclass{llncs}
\usepackage{cite}
\usepackage{multirow}
\usepackage{url}
\usepackage{graphicx}
\usepackage{mathabx}
\usepackage{subcaption}
\usepackage{geometry}
\usepackage[ ruled,lined]{algorithm2e}
\usepackage{algpseudocode}
\usepackage{color}
\SetKwInput{KwData}{Inputs}
\SetKwInput{KwResult}{Output}

\begin{document}
\title{Adaptive Learning for Service Monitoring Data}
\author{Farzana Anowar, Samira Sadaoui, Hardik Dalal
}
\institute{University of Regina, Regina, Canada, \email{anowar@uregina.ca}
\and
University of Regina, Regina, Canada, \email{sadaouis@uregina.ca}
\and
Ericsson Canada Inc., Montreal, Canada, \email{hardik.dalal@ericsson.com}
}

\maketitle

\begin{abstract}
Service monitoring applications continuously produce data to monitor their availability. Hence, it is critical to classify incoming data in real-time and accurately. For this purpose, our study develops an adaptive classification approach using Learn++ that can handle evolving data distributions. This approach sequentially predicts and updates the monitoring model with new data, gradually forgets past knowledge and identifies sudden concept drift. We employ consecutive data chunks obtained from an industrial application to evaluate the performance of the predictors incrementally. 

\end{abstract}

\section{Introduction}
\subsection{Problem and Motivation}
Industrial applications generate an enormous quantity of data, and most of the time, data must be processed and examined in real-time. Our study focuses on a real-world service monitoring and availability application. Generally speaking, monitoring services allow us to view what our servers are doing in real-time and spot patterns, spikes, and anomalies. For example, a service may be available during the week but unavailable on weekends due to maintenance, indicating a consistent pattern. A service may surge or spike due to higher demand in the year's first half. Service monitoring necessitates gathering metrics and logs from many services running on the servers in real-time. One of the main issues with these seemingly endless streams of data is to develop lightweight models that are always ready to predict and respond to unexpected changes in the data distribution. One appealing feature of these models is their ability to incorporate new data \cite{gama2014survey}. In such cases, it is crucial to update the existing classifier to accommodate new data incrementally without compromising the performance of previous data. Nevertheless, learning new information without forgetting previous knowledge raises the so-called stability–plasticity dilemma \cite{9283136}. The latter addresses the fact that a stable classifier preserves existing knowledge; however, it does not accommodate any new information, whereas a plastic classifier learns new information but does not preserve prior knowledge. In this context, the limitations of typical machine learning algorithms (MLAs) have led to incremental learning methods. 

\medskip

Recent MLAs can process large volumes of data, but as the amount of data expands, their performances suffer due to memory constraints \cite{spruyt2014curse, van2009dimensionality}. For responding to incoming data in real-time and learning from new data rapidly, incremental learning is critical in this scenario. Incremental learning, in contrast to stationary or batch learning, collects data from the dynamic environments and analyzes them in a sequential order  \cite{skmultiflow}. In terms of time efficiency and predicted accuracy, numerous studies showed that incremental learning outperforms stationary learning \cite{9283136}, \cite{zang2014comparative}, \cite{losing2018incremental}. Incremental classifiers can reinforce their knowledge without having to retrain from the beginning, i.e., without having to access the previous training data, which saves run-time greatly, and learn additional information from new data \cite{zang2014comparative}. However, data distribution may change over time in non-stationary environments, resulting in the phenomena of concept drift \cite{gama2014survey}. Adapting to concept drift may be the natural extension to incremental learning. Adaptive learning algorithms are considered as advanced incremental learning algorithms that are capable of adapting to the change in the data distribution over time \cite{gama2014survey}, \cite{skmultiflow}.

\subsection{Contributions}
Regarding our specific application from the industrial partner, we previously developed six robust labeled time-series Service Monitoring sub-datasets (called chunks) \cite{delta22}. The chunks were built using data gathered from multiple sub-servers in 2020. Each sub-server collected data every 15 seconds for 39 days. A chunk contains information of many services for one week. For our study, these chunks will be used as the incoming data by keeping in mind that the underlying data distribution may change over time.

\medskip

Since service monitoring is a continuous process, we must choose a classification algorithm that can be trained incrementally and predict incoming chunks accurately. We develop an adaptive method using Learn++. We choose Learn++ since it supports the partial-fitting mechanism to fit the model partially with new data and also handles different data distributions \cite{skmultiflow}, \cite{polikar2001learn++}. Learn++ is an ensemble of weak classifiers that produces multiple hypotheses. The advantage of using weak classifiers in an ensemble is that it reduces the training time and chances of the model over-fitting. Additionally, Learn++ refrains the base learner from undergoing catastrophic forgetting by learning incrementally. We select  K-Nearest Neighbor (KNN) as the base learner. 

\medskip

We first build the initial classifier with the first chunk, ensuring that the model has enough information to identify unseen data correctly. When the next incremental chunk becomes available, we predict each instance from that chunk and then update the model with ground truth so that the classifier can accurately predict the following instance. We select AUC and F1-score to evaluate the detection rate of the adaptive classifiers. Furthermore, we utilize False Negative Rate (FNR) as the misclassification rate to identify wrongly classified data. Performance evaluation is difficult in online learning since it is examined at different periods to check for long-term improvements. The experimental results demonstrate that the proposed method provides high detection rates (F1-score and AUC) and low misclassification rate (FNR) for the first four adaptive chunks, indicating that the adaptive model correctly classifies these chunks. On the other hand, the performance is noticeably lower for adaptive chunk\#5, showing the possibility of sudden concept drift occurrence. 

\medskip

We structure the paper as follows. Section \ref{rw5} discusses recent research on Learn++. Section \ref{data5} describes the training chunks. Section \ref{batch5} presents the Learn++ algorithm, and our proposed incremental learning framework with the `partial-fitting' mechanism. Section \ref{exp5} performs several experiments to show the efficacy of the proposed adaptive Learn++ framework. Section \ref{discuss06} discusses the cluster distribution of the last adaptive chunk, as well as possible explanations for poor performance.
Section \ref{con5} concludes with some findings and future works.
 
\section{Related Work}\label{rw5}
The authors in \cite{marwala2012line} utilized the Learn++ algorithm for the faulty gas condition monitoring application. To show the efficacy of Learn++, the authors performed two experiments. The first one showed that Learn++ learned from new data effectively, and the second one demonstrated that Learn++ accommodated with new classes efficiently. For the first experiment, a dataset with ten features was divided into training (1500 instances) and validation sub-sets (4000 instances). The training set was further divided into five sub-datasets, and as a weak-learner, MLP was utilized. The experimental results provided gradually increasing prediction accuracy. The authors utilized a training sub-set of 1,000 instances and validation sub-sets of 2,000 instances for the second experiment. The training sub-set was further divided into five sub-datasets. During the third training session, unknown classes were introduced. The classification performances did not go down for any subsequent sub-set, and accuracy gradually increased from 60\% to 95.3\%.

\medskip

A Learn++ tracker was proposed in \cite{zheng2014learn++} for robust and long-term object tracking in a non-stationary environment. The tracker dynamically maintains a set of weak classifiers trained sequentially and preserves the previously acquired knowledge. In object tracking, the distribution of samples changes a lot due to the deformation of the objects and changes in the background, especially during the transition between different sub-problems. Various subsets of weak classifiers are specified to solve different sub-problems where each classifier competes with others to be selected to suit the present environment. The authors claimed that the tracker was the first tracking method that could build an explicit model for each sub-problem, and the models could automatically be altered according to the environment. The experimental results showed that the proposed method has obtained satisfactory performances in terms of AUC. 

\medskip

To handle a large amount of data, the authors in \cite{chefrour2019novel} combined the incremental SVM (ISVM) with Learn++ (ISVM-Learn++). As a base classifier, MLP was utilized in Learn++. The two incremental classifiers were merged in two steps: 1) by parallel combination and 2) weighted sum combination. The first phase trains the two classifiers simultaneously based on a batch of data. The authors ensured no interaction between the two classifiers during training, and their design was fixed. In the second phase, the predictions from the individual classifiers are combined using the weighted sum method. Three datasets were employed; each was divided into two training sets and one test set. The experiments demonstrated that the hybrid method provided promising recognition performances and error rates. 

\medskip


In \cite{ali2021data}, the authors proposed to use Learn++ algorithm for the fault detection in Photovoltaic (PV) arrays. A simple feed-forward Multi-Layer Perceptron (MLP) with one hidden layer was utilized as the base estimator. For the experiments, the authors employed a small dataset of 600 instances with 7 features where the dataset was divided into train and test sets with 300 instances each. The train set was then subdivided into three subsets ($S_1$, $S_2$, $S_3$), with each subset being divided equally into training and testing subsets. Then, for each subset, ten weak learners were used to create a stronger learner. After the end of three training sessions with $S_1$,  $S_1 \cup S_2$, $S_1\cup S_2\cup S_3$, the test set was evaluated, and obtained an accuracy of 93\%, 92\%, and 97\%.

\section{Time-series Data Chunks} \label{data5}
For the application of the industrial partner, we constructed six time-series labeled sub-datasets (called here chunks). The data for each chunk were collected from same servers through Prometheus every 15 seconds for various services for over six consecutive weeks in the year of 2020. A chunk contains data for one week, and data is of two types: 1) Counter type  that indicates a single monotonically growing counter whose value can either increase or be reset to zero on restart, and 2) Gauge type which denotes a single numerical value that can arbitrarily go up and down \cite{ICAART2021}. An instance of a chunk indicates the service's availability, UP or DOWN, for a given point of time. For example, at 12.01 a.m., one instance may show that the service is up, but at 5.00 a.m., another instance may suggest that the same service is down. A detailed explanation of constructing the sub-datasets is provided in \cite{delta22}. Table \ref{t0501} provides a summary of the chunks that have different sizes, feature spaces and class imbalance ratios. We have a balanced ratio between two target classes for most of the chunks. The adaptive chunk\#2 has a maximum imbalanced ratio of 1:3, which is still a reasonable and acceptable imbalance ratio. As a result, no sampling strategy is employed to balance this chunk.

\begin{table}[h!]
\centering
\renewcommand{\arraystretch}{1.3}
\caption{Weekly Chunks and Labels}
\label{t0501}
\begin{tabular}{|c|c|c|c|c|c|}
\hline
\textbf{Chunk} & \textbf{Size} & \textbf{Dimensionality} &\textbf{Class 0} & \textbf{Class 1}  &\textbf{Imbalanced Ratio} \\ \hline
Initial      & 9099   &75 &   3733               &    5366     & $\approx$ 1:1.5        \\ \hline 
Adaptive\#1            & 10080 &100 &   4310               &     5770        & $\approx$ 1:1     \\ \hline
Adaptive\#2             & 10080  &  117 &   2313               &  7767          & $\approx$ 1:3 \\ \hline
Adaptive\#3             & 10080 &  76 &  4525                &     5555         &  $\approx$ 1:1\\ \hline
Adaptive\#4             & 10080 &   100 &   5556               &    4524           & $\approx$ 1:1\\ \hline
Adaptive\#5             & 4534 &    120 &    2306              &     2228         & $\approx$ 1:1\\ \hline

\end{tabular}
\end{table}

\medskip

However, the chunks come with different dimensionality that needs to be reduced to be able to perform the learn++ algorithm. To do so, we utilize the total explained variance ratio (TEVR) metric to provide each feature's amount of information. The initial chunk has a total TEVR of 99.9920\% for 75 features. Table \ref{variance05} presents the TEVR for all the adaptive chunks for the first five features. For instance, for adaptive chunks \#1 and \#2, the first feature itself contains $\approx$98.87\% and $\approx$99.24\% of information, respectively. It is evident that we do not lose much information even after first five features. As a result, we choose the dimensionality of 75 (the first 75 features among all the features) for all the adaptive chunks. 

\begin{table}[!h]
    \caption{TEVR of First Five PCs for Each Adaptive Chunk}
    \label{variance05}
    \renewcommand{\arraystretch}{1.2}
    \resizebox{\textwidth}{!}{%

    \begin{minipage}{.5\linewidth}
      \centering
       \begin{tabular}{|c|c|c|}
\hline
\textbf{Chunk}      & \textbf{Feature} & \textbf{\begin{tabular}[c]{@{}c@{}}TEVR \end{tabular}} \\ \hline
\multirow{5}{*}{1} & 1                  & 0.98879                                 \\ \cline{2-3} 
                   & 2                  & 0.00107                                  \\ \cline{2-3} 
                   & 3                  & 0.00077                                 \\ \cline{2-3} 
                   & 4                  & 0.00066                                 \\ \cline{2-3} 
                   & 5                  & 0.00059                                \\ \hline 
\multirow{5}{*}{2} & 1                  & 0.99246                                 \\ \cline{2-3} 
                   & 2                  & 0.00747                                 \\ \cline{2-3} 
                   & 3                  & 4.73E-05                                    \\ \cline{2-3} 
                   & 4                  & 2.65E-06                                    \\ \cline{2-3} 
                   & 5                  & 2.43E-06                                    \\ \hline
                   
\multirow{5}{*}{3} & 1                  & 0.99947                                 \\ \cline{2-3} 
                   & 2                  & 0.00046                                 \\ \cline{2-3} 
                   & 3                  & 5.79E-06                                    \\ \cline{2-3} 
                   & 4                  & 5.31E-06                                    \\ \cline{2-3} 
                   & 5                  & 4.45E-06                                    \\ \hline
    \end{tabular}
    \end{minipage}%
    \begin{minipage}{.5\linewidth}
\centering
\begin{tabular}{|c|c|c|}
\hline
\textbf{Chunk}      & \textbf{Feature} & \textbf{\begin{tabular}[c]{@{}c@{}}TEVR \end{tabular}} \\ \hline
        
\multirow{5}{*}{4} & 1                  & 0.98718                                 \\ \cline{2-3} 
                   & 2                  & 0.00144                                 \\ \cline{2-3} 
                   & 3                  & 0.00116                                 \\ \cline{2-3} 
                   & 4                  & 0.00107                                 \\ \cline{2-3} 
                   & 5                  & 0.00067                                 \\ \hline
\multirow{5}{*}{5} & 1                  & 0.98103                                 \\ \cline{2-3} 
                   & 2                  & 0.00237                                 \\ \cline{2-3} 
                   & 3                  & 0.00227                                 \\ \cline{2-3} 
                   & 4                  & 0.00119                                 \\ \cline{2-3} 
                   & 5                  & 0.00103                                 \\ \hline
\end{tabular}
    \end{minipage} }
\end{table}

\section{Adaptive Learning}\label{batch5}
\subsection{An Overview}
Change in data distribution can have a significant impact on the predictive performances. Decision models operating in dynamic environments require adapting to data over time; otherwise, their performance can be misleading \cite{gama2014survey}. Adaptive models predict and learn from evolving data with unknown dynamics \cite{gama2014survey}. They update themselves continuously to  changes without going through the entire time-consuming retraining process \cite{skmultiflow} and still retain the previous knowledge in the model's memory, boosting accuracy and lowering the risk of model's failure \cite{skmultiflow}. An adaptive learning procedure works as following \cite{gama2014survey}, \cite{zhang2021adaptive}:

\begin{itemize}
    \item Predict: When new data ($x_i$) arrives, a prediction ($\hat{y_i}$) is made using the current classifier ($C_i$).
    
    \smallskip
    
    \item Diagnose: When true label ($y_i$) is received, estimate the loss between $y_i$ and $\hat{y_i}$. 
    
    \smallskip
    
    \item Update: ($x_i, y_i$) is used to update $C_i$ to obtain $C_{i+1}$. While updating, the loss estimation is utilized to identify if any distribution change has occurred from the distribution of previous data.

\end{itemize}


\medskip

Adaptive learning is beneficial for large training datasets that cannot fit into the memory of a single machine and also for continuous data streams. In short, it is capable of processing an infinite data stream with finite resources in terms of processor and memory. 


\medskip

Learning under changing data distributions necessitates not only updating the model with new information but also forgetting old information \cite{gama2014survey}. The literature offers several adaptive meta-learning (ensemble) methods by keeping multiple learners in the ensemble and using the partial-fitting mechanism \cite{skmultiflow}. For this research, we adopt Learn++ \cite{polikar2001learn++}, \cite{1007781}, an adaptive method to update the classifier gradually with the time-series chunks and retain past knowledge using the Scikit-Multiflow setting. Learn++ can also accommodate new data distributions that may be introduced with new target classes \cite{polikar2001learn++}. Learn++ algorithm employs a number of weak classifiers to generate a stronger classifier that predicts an instance by updating its weights. For this approach, a window of incoming data is first collected and then used to build adaptive models. During the adaptive learning step, Learn++ does not require access to previously used data, and at the same time, it does not forget immediately what it has already learned; rather forgets gradually by keeping the weak learners in the ensemble. Thus, it effectively addresses the stability and plasticity issue. Hence, Learn++ is memory efficient. Due to these advantages, this approach is efficient for big data and when real-time responses are vital, like our service monitoring application. 

\subsection{Learn++ Algorithm}
Learn++ takes advantage of the ensemble method to learn from new data \cite{skmultiflow}, \cite{1007781}. An ensemble approach creates weak classifiers and combines them to generate a much stronger meta-learner to improve the performances \cite{dogan2019weighted}. Usually, any ensemble method is more accurate than a single classifier \cite{1007781}, \cite{anowar2019multi}. A dataset is first divided into several sub-datasets. Learn++ builds several weak classifiers for each sub-dataset, whose outputs are then combined through the weighted majority voting to obtain the composite prediction for that sub-dataset. Later, the weights are updated dynamically for the misclassified data. The latter are then included in the next sub-dataset, where the weak classifier emphasizes these incorrectly predicted data.  

\medskip

In this way, the new model always has the information from the previous data. In Learn++, the weak classifiers are addressed as weak hypotheses. A weak classifier can be any supervised classifier that can classify at least half of the data correctly from a dataset \cite{1007781}. Additionally, the weak classifiers have the advantage of rapid training. However, unlike stronger classifiers, they generate a rough approximation of the final classification decision, which allows avoiding a significant amount of training time \cite{1007781}. On the other hand, a strong classifier spends much time fine-tuning the decision boundary. Since a weak classifier eliminates this costly fine-tuning, it allows faster training and less chance of model over-fitting, which is a much-desired feature for any classifier \cite{1007781}. The main steps of Learn++ are given below:

\medskip

Learn++ first initializes equal weight (1/$i$, \textit{i} is total number of data) distribution for all data. Then, it divides a sub-dataset into training and testing sub-sets randomly. The equal weight distribution gives equal probability to each instance to be selected for training. A weak classifier is then trained, and tested on both training and testing sub-sets to obtain $Hypothesis_t$. The error is calculated by summing the distributed weights for the misclassified data, as follows \cite{polikar2001learn++}, \cite{1007781}: 
\begin{equation}
    e_t = \sum_{i:Hypothesis_t (x_i \neq y_i)}  D_t (i)= W_1 + W_2 + ... + W_i
\end{equation}

\noindent where $t$ is the number of weak classifiers to be generated for each sub-dataset, $e_t$ is the total error of the misclassified data, $W_1, W_2, ... W_i$ are the weights of each misclassified data by the weak classifiers, $x_i$ is a data, and $y_i$ the class label of $x_i$. Learn++ includes one condition is that $e_t$ must be less than 0.5. If $e_t$ $<$ 0.5, the error is normalized (see Equation \ref{nError}), otherwise the current $Hypothesis_t$ is discarded, and a new training sub-set is selected \cite{polikar2001learn++}. 

\begin{equation}
\label{nError}
    \beta_t = \frac{e_t}{1- e_t}, ~ where~  0< \beta <1
\end{equation}

The above process continues until the specified number of weak classifiers is attained. Next, all the hypotheses generated are combined using the weighted majority voting to provide the composite hypothesis ($H_t$). The voting weights of $H_t$ are computed as the logarithms of the reciprocals of $\beta_t$ as shown in Equation \ref{log} \cite{polikar2001learn++}. Therefore, the hypotheses that perform well on their training and test sub-sets are given higher voting powers. A classification decision is then taken based on the combined performance of each hypothesis, which constitutes $H_t$. $H_t$ decides the target class that receives the maximum vote from all \textit{t} hypotheses.

\begin{equation}
\label{log}
    H_t = arg~max_{y \in Y} \sum_{t:Hypothesis_t(x)=y} log \frac{1}{\beta_t}
\end{equation}

\noindent where $Y$ is the set of class labels. The composite error ($E_t$) of $H_t$ is computed as the sum of misclassified data's weights as shown in Equation \ref{22}. Next, the normalized composite error ($B_t$) is computed as demonstrated in Equation \ref{NE} \cite{polikar2001learn++}. 

\begin{equation}
\label{22}
    E_t = \sum_{i:H_t (x_i \neq y_i)} D_t (i) 
\end{equation}

\begin{equation}
\label{NE}
    B_t = \frac{E_t}{1-E_t}, 0 < B_t < 1
\end{equation}

$B_t$ is used to update the distribution of weights as shown below. According to this rule, if a data is correctly classified by the composite hypothesis, its weight is decreased by multiplied with $B_t$ ( $B_t$ is always less than 1). If an instance is misclassified, the distribution weight remains unchanged \cite{polikar2001learn++}. Thus, this weight updation reduces the probability of correctly classified instances being chosen into the next traininig sub-dataset, while increasing the probability of misclassified instances to be selected \cite{polikar2001learn++}. Therefore, the algorithm focuses on the misclassified data. In addition to that, if all data are correctly classified by a weak classifier, the Learn++ stops adding the weak classifier in an ensemble. 

  \[
    w_{t+1} (i)= w_t (i) \times \left\{
                \begin{array}{ll}
                  B_t, ~ if ~ H_t (x_i)=y_i\\
                  1, otherwise
                \end{array}
              \right.
  \]

\noindent where $w_{t+1} (i)$ is the new weight of data $i$ for $t+1$-th weak classifier, $w_t$ is the old weight of $t$-th weak classifier. For Learn++, weight updation is the heart of this algorithm and this makes it different from other ensemble methods. With the updated weights, Learn++ emphasizes on the wrongly classified data, and increase the probability of the misclassified data of being selected to the next training sub-dataset \cite{1007781}. In other words, the algorithm is forced to focus more on instances that are hard to classify.


\subsection{Proposed Adaptive Setting}\label{bic}
Learn++ in \cite{polikar2001learn++} divides the dataset into training and testing sets randomly based on the weight distribution. However, a random distribution ratio may not be practical for challenging real-world data like ours, and as a result, the classification performance can be significantly lower. Hence, instead of using training and testing sets, we propose to build a steady pre-trained model first by training the Learn++ classifier with the initial chunk and optimizing the hyper-parameters of Learn++ and the KNN classifier. Second, whenever the next chunk becomes available, we do not partition the chunk into training and testing sets; rather, the devised approach incrementally classifies each new chunk based on this pre-trained model. After predicting each data from the chunk, the classifier gets updated with ground truth. At the same time, if the prediction is made wrong, the weight gets updated to put more effort into learning the misclassified data. While updating the classifier, Learn++ does not re-train from scratch; instead, it adjusts the current classifier with ground truth while remembering the previous knowledge. Thus, the adaptive model is constantly updated with the newly added data with correct information, allowing more accurate prediction of the next unseen data.

\begin{figure*}[!h]
\centering
  \includegraphics[width=11cm, height=8cm]{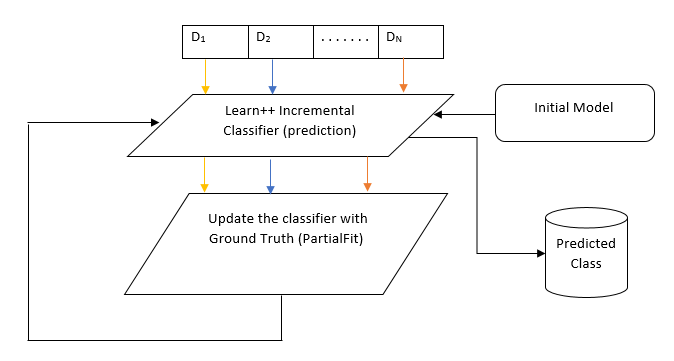}
  \caption{Learn++ Classifier with Updated Adaptive Classifier}
  \label{pf}
\end{figure*}

\medskip

Using our approach, we determine the weight of each data using the prediction result rather than the prediction of each sub-dataset. In this way, we ensure that we pay equal attention to each data, and each data gets equal importance while updating a classifier. This strategy is vital for our service monitoring application since data are challenging. Furthermore, unlike any chunk-based approach, our proposed method keeps the model always lightweight and allows it to exit or stop the adaptive process at any time. After predicting the unseen data, we store each predicted result to compare the ground truth and the predicted class for each adaptive chunk and calculate the performance of each adaptive chunk as a whole.

\medskip

We additionally evaluate the performances of the adaptive chunks with three performance metrics (F1-score, AUC, and FNR). High F1-score and AUC and low FNR indicate that the devised classifier is classifying the chunks accurately. We also keep track of the correctly classified data of each adaptive chunk so that we can have count of total correctly classified data as shown in Table \ref{correct}. In Algorithm \ref{algo5}, we present the steps of devised adaptive learning approach with Learn++. 

\begin{algorithm}[!h]
\caption{Learn++ adaptive Classifier on Service Monitoring Chunks (instance by instance)}
\label{algo5}

\KwData{$ initChunk, incChunk_i$, PC}
\KwResult{$incrementalModel_i$}

1. \For{$IncChunk_i$}{
        \textcolor{blue}{ /*Get same dimension as initChunk*/}
        
         1.1. $reducedIncChunk_i$ $\gets$ Slice ($incChunk_i$, PC)\;
        }

2. Select and configure \textbf{\textit{baseEstimator}}\; 
3. Set \textbf{\textit{windowSize}}, \textbf{\textit{numberEstimators}}\;

4. LPP $\gets$ configure with \textit{BaseEstimator, windowSize, numberEstimators}\;

\textcolor{blue}{/*Build Initial Classifier for Pre-training only*/}

5. initModel $\gets$ Train($LPP$, $initChunk$)\;

6. $incrementalModel_{0}$ = initModel\;
\textcolor{blue}{/*Incremental Learning for Each adaptive Chunk*/}

7. \For{each $reducedIncChunk_i$}{
7.1. correct\_cnt = 0\;
7.2. $Y\_PRED_i$=[]\;

7.3.  \For{each $instance_j$ from $reducedIncChunk_i$}
    {

 \textcolor{blue}{/*Predict class for an unseen data*/}
 
 7.3.1. $y\_pred_j$ $\gets$\
\textbf{predict}($incrementalModel_{i-1}$, $instance_j$)\; 
7.3.2. $Y\_PRED_i$.append($y\_pred[j]$)\;
7.3.3. \If { y[j] == y\_pred[j]}{
        correct\_cnt += 1\;}
        
 \textcolor{blue}{/*incrementally update the model with Ground Truth*/}
 
7.3.4. $incrementalModel_{i}$ $\gets$ partial\_fit($incrementalModel_{i-1}$, $instance_j$)  \;

} 
 
\textcolor{blue}{/*Run evaluation for $reducedIncChunk_i$*/}

7.4. Evaluate performance ($y_i$, $Y\_PRED_i$)\;

7.5. Return $incrementalModel_i$\;}
\end{algorithm}

\medskip
Figure \ref{pf} demonstrates how Learn++ is getting adjusted with new data after making a prediction. The adaptive classifiers in Scikit-Multiflow are specifically designed in a way that they are robust to concept drift \cite{skmultiflow}. In the presence of concept drift, the new data are more hard to learn meaning that the concept drift occurs and the predictive performance of new data substantially decreases \cite{zhang2021adaptive}. 

\medskip

In Scikit-multiflow adaptive setting, we need to tune four hyper-parameters for Learn++ algorithm: base\_estimator (KNN), window\_size (chunk's size), and n\_estimators (number of estimators due to memory limitation), error\_threshold (Table \ref{tbatch005}). We select 3 estimators for each ensemble to make the whole incremental process less computationally expensive as we have high number of data in each chunk. Here, window\_size can receive different sizes of chunks sequentially. 

\begin{table}[h!]
\centering

\caption{Hyper-parameters for Learn++ Classifier}
\label{tbatch005}
\renewcommand{\arraystretch}{1.3}
\begin{tabular}{|c|c|c|}
\hline
\textbf{Hyper-parameter} & \textbf{Explanation}                 & \textbf{\begin{tabular}[c]{@{}c@{}}Chosen\end{tabular}}   \\ \hline
base\_estimator           & any weak classifier             & KNN                                                                     \\ \hline
window\_size              & variable          & \begin{tabular}[c]{@{}c@{}}size of each chunk\end{tabular} \\ \hline
n\_estimators             & number of classifiers per ensemble & 3                                                                      \\ \hline
error\_threshold             & \begin{tabular}[c]{@{}c@{}} to keep the classifier that has\\ error smaller than threshold\end{tabular} & 0.5 \\ \hline
\end{tabular}%
\end{table}
\section{Experiments}\label{exp5}

Finding the best combination of Learn++ and KNN was a challenging task for us. We tried several combinations of all the hyper-parameters of Learn++ and KNN using the initial chunk. We utilize the optimal configuration for all the adaptive chunks.  
\subsection{Performance of Initial Classifier}\label{init5}
We first build the initial classifier using the initial chunk. We choose KNN classifier as the base-estimator over other MLAs since: 1) it is simple but proved to be an effective classifier \cite{abu2019effects}, 2) it has few hyper-parameters to tune, hence it is faster, 3) it deals efficiently with noisy data, and 4) it can be used for multi-class problem. KNN first calculates the distance from the test data to all other data in a dataset. Then, it selects it's $k$ nearest neighbors, assigns a target class label for the test data that is the most frequent in the nearest neighbors \cite{abu2019effects}. We also tried using other classifiers such as MLP. However, since the service monitoring chunks contain comparatively large amount of data, and MLP requires higher number of hyper-parameters to configure, therefore, Learn++ takes a huge amount of time to categorize the incoming chunks. To have less computation complexity, we consider nearest neighbor number of 3. When using a value greater than 3, the computation time increases significantly, at the same time the performance remains approximately the same. We also utilize distance = `minkowski' for KNN with $p=2$. $p$ is the power parameter for the distance metric. If $p = 1$, this is equivalent to using manhattan distance, and if $p=2$, this is equivalent to euclidean distance \cite{scikit-learn}. 

\medskip

We choose F1-score, AUC, and FNR to evaluate the performances. F1-score is the harmonic mean of precision and recall. A higher F1-score means higher detection rate in terms of precision and recall \cite{anowar2019multi}. AUC tells how much an adaptive model is capable of distinguishing between the classes. Higher the AUC, the better the model is at predicting the target classes \cite{jimenez2012insights}. FNR presents the misclassification rate of a classifier, and lower the FNR, better the classifier is \cite{anowar2019multi}. Table \ref{t2-05} presents the performance of the initial classifier. We obtain the F1-score, AUC of 99.72\% and  99.68\% respectively, FNR of 0.20\%. This higher performance of the initial classifier will help the adaptive classifier to predict the unseen data correctly, and boost up the performances of the adaptive chunks. 

\begin{table}[h!]
\centering
\renewcommand{\arraystretch}{1.3}
\caption{Performance of Initial Classifier }
\label{t2-05}
\begin{tabular}{|c|c|c|c|c|c|}
\hline
\textbf{Chunk} & \textbf{PCs}    & \textbf{F1-score} & \textbf{\begin{tabular}[c]{@{}c@{}}AUC\end{tabular}} & \textbf{FNR}  \\ \hline
Initial              & 75                                                                    & 0.9972             &   0.9968      &  0.0020                                                                                                                                                                                    \\ \hline
\end{tabular}%
\end{table}

\subsection{Performance of Adaptive Classifiers}
Table \ref{t3-05} demonstrates the performances of the adaptive chunks. For adaptive chunk\#1, we attain F1-score, and AUC of 98.75\%, and 98.76\% respectively, and FNR of 0.28\%. We achieve the highest F1-score, AUC of 99.86\%, 99.80\% respectively with chunk\#3, and lowest FNR of 0.0.06\%  with chunk\#4. However, the adaptive chunk\#5 returns the lowest performances across all three metrics. It provides 64.64\%, and 60.46\% F1-score and AUC respectively, and 68.01\% of FNR. However, FNR gradually decreases from adaptive chunk\#1 to \#4 from 0.28\% to 0.06\%. Except the adaptive chunk\#5, the other four adaptive chunks provide satisfactory detection, and misclassification rate. Here, for chunk\#5, the adaptive classifier is having difficulties to learn new data, and performance has gone down significantly.

\begin{table}[h!]
\centering
\caption{Performance of Adaptive Classifiers }
\label{t3-05}
\renewcommand{\arraystretch}{1.3}
\begin{tabular}{|c|c|c|c|c|c|c|}
\hline
\textbf{\begin{tabular}[c]{@{}c@{}}Adaptive\\ Chunk\end{tabular}} & \textbf{F1-score} & \textbf{\begin{tabular}[c]{@{}c@{}}AUC\end{tabular}} & \textbf{FNR}  \\ \hline
 
\#1                                  & 0.9875             &     0.9876     &    0.0028                                                                                                                      \\ \hline
\#2                                & 0.9665             &     0.9006      &    0.0016                                                          \\ \hline                                                           
\#3                                     & 0.9986                 &      0.9980        & 0.0010                                                                \\ \hline                                                       
\#4                                 & 0.9738             &       0.9776        &    0.0006                                                       \\ \hline                                                        
\#5                                   & 0.6464             &     0.6046      &    0.6801                                                                                                                   \\ \hline
\end{tabular}%
\end{table}


\medskip

Furthermore, in Table \ref{correct}, we present the number of correctly classified Service Monitoring data for each adaptive chunk by the devised Learn++ algorithm. For the adaptive chunk\#1, Learn++ correctly classified 98.59\% (9,938 data) of data out of 9,099 data, whereas, for adaptive chunk\#2, the number of accurately classified data goes down by 388. The adaptive algorithm correctly identify the maximum 10,065 data, and misclassify only 15 data from the adaptive chunk\#3. From adaptive chunk\#4, 9839 data are correctly classified. However, out of 4534 data for adaptive chunk\#5, only 2731 data have been classified accurately. From our previous study \cite{delta22}, we notice that we attain the lowest trustworthiness, and similarity of 59.91\% and 71.61\% for the adaptive chunk\#5.

\begin{table}[!h]
\centering
\caption{Correctly Classified Data for Incoming Chunks}
\label{correct}
\renewcommand{\arraystretch}{1.3}
\begin{tabular}{|c|c|c|c|}
\hline
\textbf{\#Chunk} & \textbf{\begin{tabular}[c]{@{}c@{}}Correctly Classified\\ Data\end{tabular}} & \textbf{\begin{tabular}[c]{@{}c@{}}Incorrectly Classified\\ Data\end{tabular}} & \textbf{\begin{tabular}[c]{@{}c@{}}Percentage of \\ Correctly Classified \\ Data\end{tabular}} \\ \hline

\#1              & 9938          &    142                                                           & 98.59\%                                                                                          \\ \hline
\#2              & 9550       &       530                                                           & 94.74\%                                                                                          \\ \hline
\#3              & 10065    &      15                                                              & 99.85\%                                                                                            \\ \hline
\#4              & 9839     &     241                                                               & 97.60\%                                                                                          \\ \hline
\#5              & 2731     &    1803                                                                & 60.23\%                                                                                          \\ \hline
\end{tabular}
\end{table}

\section{Discussion}\label{discuss06}
\begin{figure*}[h!]
\centering
    \begin{subfigure}[b] {0.25\textwidth}
        \centering
        \includegraphics[width=\textwidth]{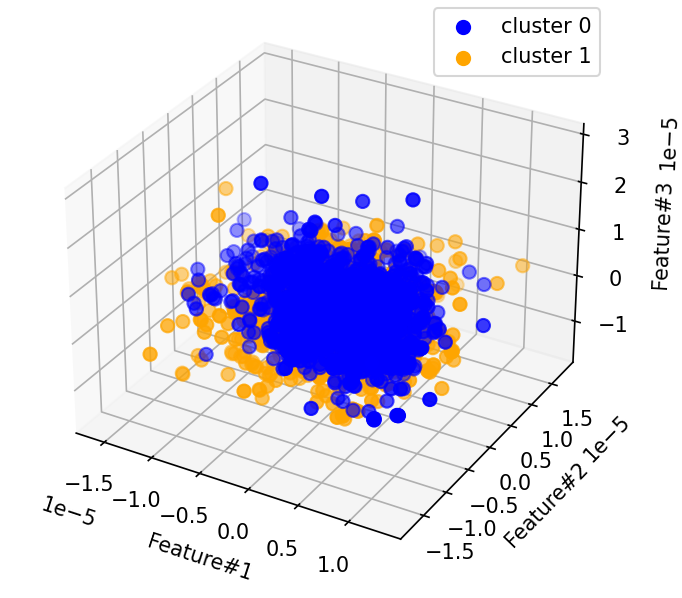}
        \caption{Adaptive Chunk\#5}
    \end{subfigure}
    \quad
    \begin{subfigure}[b]{0.25\textwidth}  
        \centering 
        \includegraphics[width=\textwidth]{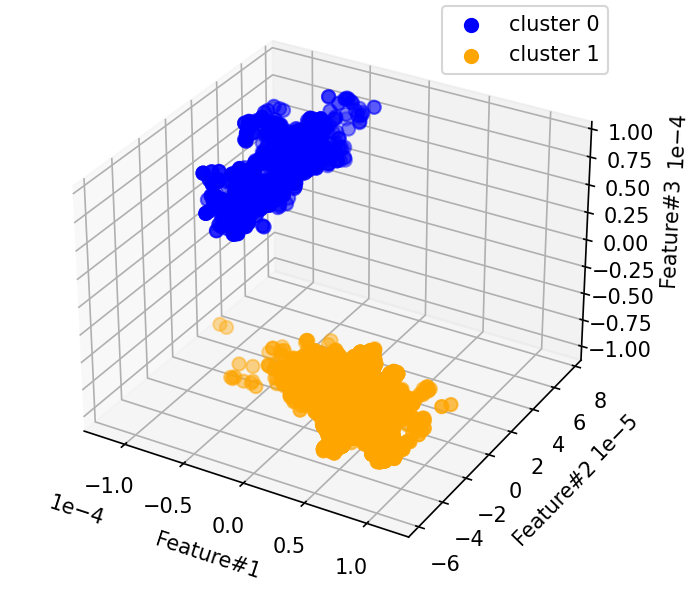}
        \caption{Adaptive Chunk\#3}
    \end{subfigure}
    \caption{Cluster Distribution }
    \label{clus06}
\end{figure*}

Concept drift primarily refers to changes in data distribution where the relation between the input data and target variable changes over time \cite{iwashita2018overview}. However, this change may occur suddenly, incrementally, or gradually \cite{iwashita2018overview}. Because of the lower performance of adaptive chunk\#5, we may conclude that a sudden drift happened. In contrast, gradual drift is present for the first four chunks since the performance goes up and down with a  small gap. Also, outliers are not considered concept drifts \cite{gama2014survey}. In Figure \ref{clus06}-(a), we present the clustering distribution with 2 clusters for adaptive chunk\#5, where a good amount of data are far away from regular data points from both the clusters. This indicates that there is a high possibility of having outliers. We also show the cluster distribution of the third adaptive chunk in Figure \ref{clus06}-(b) for which we attain the best detection performance, and the two clusters are separated from each other, and no significant amount of distant data are visible.

\section{Conclusions and Future Works}\label{con5}
Service monitoring is critical because it allows us to take proactive measures when problems arise, ensure data security, limit server downtime, and promptly resolve problems. Since service monitoring allows us to ensure the best usage of resources, it is essential to classify data in real-time. As a result, we proposed a Learn++ adaptive classifier framework to handle the service monitoring data efficiently in dynamic environment. Under the hood of Learn++, as the base estimator, KNN was utilized. We tried several combinations between Learn++ and KNN to find the optimal topology, which yielded good detection and misclassification rate on adaptive chunks, except chunk \#5 that may suffer from sudden concept drift and outliers.

\medskip

This current work leads to several future research directions for the same industrial application. We intend to thoroughly investigate different types of concept drift and several concept drift detection methods and choose the most suitable methods for handling our Service Monitoring application. We are also interested in incorporating the clustering-based outlier detection to increase the performance further, especially for adaptive chunk\#5. 
 
\section{Acknowledgement}
We would like to express our gratitude to Global AI Accelerator (GAIA) Ericsson, Montreal for collaborating with us for this research work, and the Observability team for allowing us to access the data. This work was funded and supported by GAIA and Mitacs Accelerate grant (IT16751)

\bibliographystyle{splncs}
\bibliography{ref} 

\end{document}